# Modeling of Social Transitions Using Intelligent Systems

Hamed .O.Ghaffari[*], Witold Pedrycz[†] & Mostafa.Sharifzadeh[*]
*Department of Mining and Metallurgical Engineering,*
*Amirkabir University of   Technology, Tehran, Iran*
h.o.ghaffari@gmail.com sharifzadeh@aut.ac.ir

[†]*Department of Electrical and Computer Engineering ,University of Alberta*
*Alberta, Canada - pedrycz@ece.ualberta.ca*

*Abstract*
*In this study, we reproduce two new hybrid intelligent systems, involve three prominent intelligent computing and approximate reasoning methods: Self Organizing feature Map (SOM), Neruo-Fuzzy Inference System and Rough Set Theory (RST),called SONFIS and SORST. We show how our algorithms can be construed as a linkage of government-society interactions, where government catches various states of behaviors: "solid (absolute) or flexible". So, transition of society, by changing of connectivity parameters (noise) from order to disorder is inferred.*

## 1. Introduction

Complex systems are often congruous with uncertainty and order-disorder transitions. Apart of uncertainty, fluctuations forces due to competition of between constructive particles of system drive the system towards order and disorder. There are prominent examples which their behaviors show such anomalies in their evolution, i.e., physical systems, biological and financial systems [1]. In other view, in monitoring of most complex systems, there are some generic challenges for example sparse essence, conflicts in different levels, inaccuracy and limitation of measurements ,which in beyond of inherent feature of  such interacted systems are real obstacles  in their analysis and predicating of behaviors. There are many methods to analyzing of systems include many particles that are acting on each other, for example statistical methods [2], Vicsek model [3] etc.

A discrete motivation in this way is finding out of "main nominations of each distinct behavior which may has overlapping, in part, to others". This advance is to bate of some mentioned difficulties that can be concluded in the "information granules" proposed by Zadeh [4]. In fact, more complex systems in their natural shape can be described in the sense of networks, which are made of connections among the units. These units are several facets of information granules as well as clusters, groups, communities, modules [5]. In this study, we reproduce two hybrid intelligent systems [9], [10]: Self Organizing Neruo-Fuzzy Inference System (SONFIS) and Self Organizing Rough Set Theory (SORST), and then investigate several levels of responses against the real information (cumulative information flow). We show how these methods can produce (mimic) the interacted behaviors of government-nation. Mutual relations between the proposed algorithms layers identify order-disorder transferring of similar systems.

## 2. Methods

In this section based upon Self Organizing feature Map (SOM) [6], adaptive Neuro-Fuzzy Inference System (NFIS) [7] and Rough Set Theory (RST) [8], we reproduce: Self Organizing Neuro-Fuzzy Inference System (SONFIS) and Self Organizing Rough Set Theory (SORST) [9], [10]. In this study our aim is to investigate order-disorder transition in the mentioned systems.  The mentioned algorithms use four basic axioms upon the balancing of the successive granules assumption:
Step (1): dividing the monitored data into groups of training and testing data

Step (2): first granulation (crisp) by SOM or other crisp granulation methods
  Step (2-1): selecting the level of granularity randomly or depend on the obtained error from the NFIS or RST (regular neuron growth)
  Step (2-2): construction of the granules (crisp).
Step (3): second granulation (fuzzy or rough granules) by NFIS or RST
  Step (3-1): crisp granules as a new data.
  Step (3-2): selecting the level of granularity; (Error level, number of rules, strength threshold, scaling of inserted information...)
  Step (3-3): checking the suitability. (Close-open iteration: referring to the real data and reinspect closed world)
  Step (3-4): construction of fuzzy/rough granules.
Step (4): extraction of knowledge rules

Balancing assumption is satisfied by the close-open iterations: this process is a guideline to balancing of crisp and sub fuzzy/rough granules by some random/regular selection of initial granules or other optimal structures and increment of supporting rules (fuzzy partitions or increasing of lower /upper approximations ), gradually.

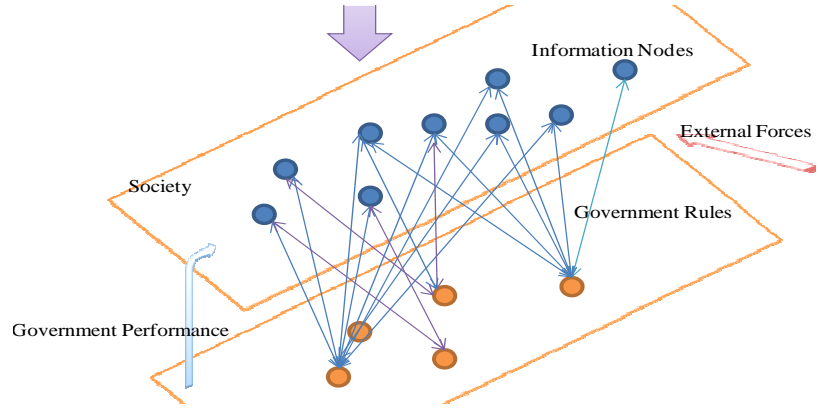

**Figure1. A general schematic of Society-Government network**

The overall schematic of SONFIS and SORST-AS has been shown in Figure 2 and Figure 3. In first granulation step, we can use a linear relation or a power-law which is given by:

$$N_{t+1} = \alpha N_t + \Delta_t ; \Delta_t = \beta E_t + \gamma \quad (1)$$

$$N_{t+1} = N_t^{\alpha} + E_t^{\beta} + \gamma \quad (2)$$

where $N_t = n_1 \times n_2 ; |n_1 - n_2| = Min.$ is number of neurons in SOM or Neuron Growth (NG); $E_t$ is the obtained error (measured error) from second granulation on the test data and coefficients must be determined, depend on the used data set. One may interpret Eq. (1) as evolution of crisp granules in time step t which are depend on the effects of the pervious state of granules ($\alpha$), the impact of the regulator (here NFIS or RST) performance ($\beta$) and other external forces ($\gamma$). In other view and using complex networks theory, we may make a hierarchical complex network in which information nodes in a society depend on their requirements (or in a compulsory way) are connected to the government's rule(s) and by a dominant law (Eq1. or Eq.2) on the relation between two sides (Figure 1). Considering of several aspects of intellectuality, uncertainty, evolution of nodes (society and/or government) and other possible fascinating results are some of the motivations of such structures.

We can assume interactions of the two layers of algorithms as the behaviors of complex systems such: society and government, where reactions of a dynamic community to an "absolute (solid-dictatorship) or flexible (democratic)" government (regulator) is controlled by correlation (noise) factors of the two simplified systems. In absolute case, the second layer (government / regulator) has limited rules with stable learning iteration for all of matters. In first layer, society selects most main structures of the stimulator where these clusters upon the reaction of government and pervious picked out structures will be adjusted. In facing this, flexible regulator has ability of adapting with the evolution of society.

This situation can be covered by two discrete alternatives: evolution of constitutive rules (policies) over time passing or a general approximation of the dominant rules on the emerged attitudes. In the latter case the legislators can considers being conflicts of the emerged states. With variations of correlation factors of the two sides (or more connected intelligent particles), one can identify point (or interval) of behavior changes of society or overall system and then controlling of society may be satisfied. The scaling of the observed information, definitely, has a great role in identification of long and short policies, especially for democratic government.

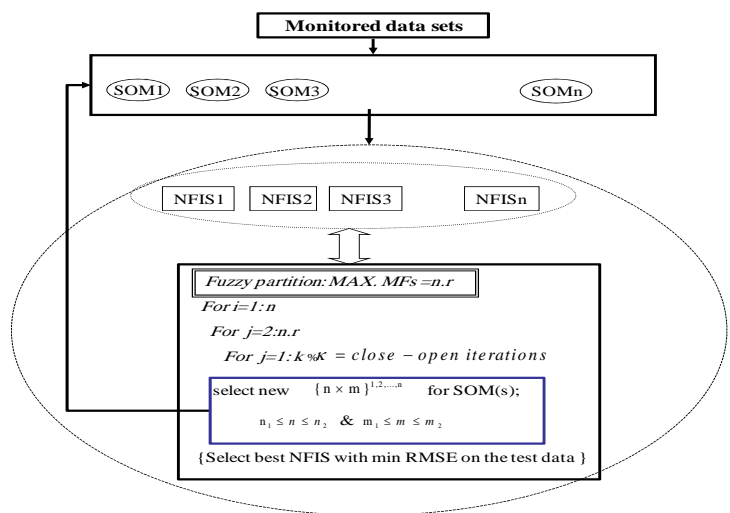

**Figure 2. Self Organizing Neuro-Fuzzy Inference System (SONFIS)**

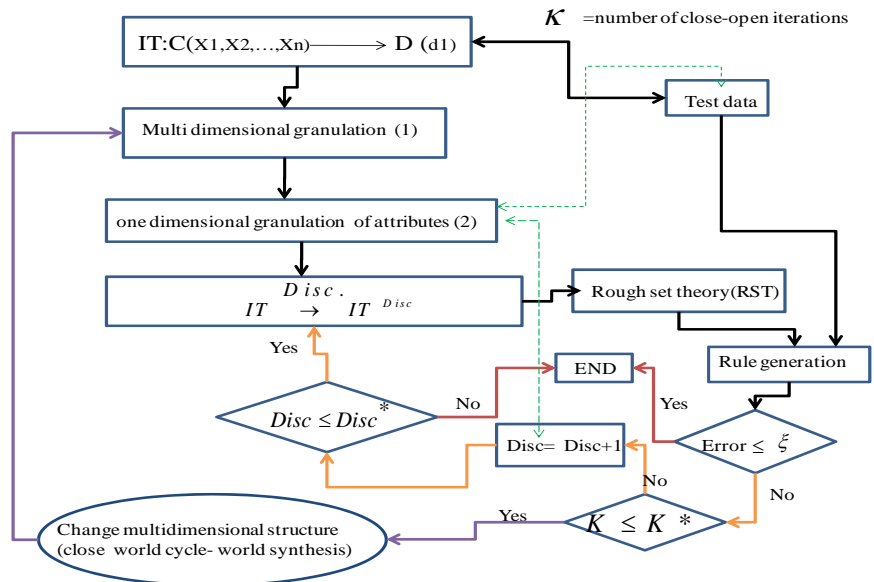

**Figure 3. Self Organizing Rough Set Theory-Adaptive Scaling (SORST-AS)**

## 3. Phase transitions: society & revolutionary state

In this part of paper, we ensue our algorithms on the "lugeon data set" [9], [10]. To evaluate the interactions, we follow two procedures where phase transition measure is upon the crisp granules (here NG): 1) second layer takes a few rules, extracted by using NFIS; 2) considering elicited rules by RST and under an approximated progress (with changing of scaling).

Analysis of first situation is started off by setting the number of close-open iteration and maximum number of rules equal to 30 and 3 in SONFIS respectively. The error measure criterion in SONFIS is Root Mean Square Error (RMSE), given as below:

$$RMSE = \sqrt{\frac{\sum_{i=1}^{m}(t_i - t_i^*)^2}{m}}, \quad (3)$$

where $t_i$ is output of SONFIS and $t_i^*$ is real answer; m is the number of test data(test objects). In the rest of paper, let m=93 and number of inserting data set =600. By employing of Eq. (1) in SONFIS and $\beta=.001$ and $\gamma=.5$; the general patterns of NG and RMSE versus time steps and variations of $\alpha$ can be observed (Figure 4). Figure 4 indicates how the neurons fluctuations with time passing reveal more chaos while the phase transition step can be transpired in $\alpha=.8-.85$. The integration of average neuron growth, in a bar graph, exhibits the phase transferring interval (Figure 5). We can reap other interesting results when the numbers of rules are added and in this way, the government persists on the same number of learning (i.e., semi-absolute government).

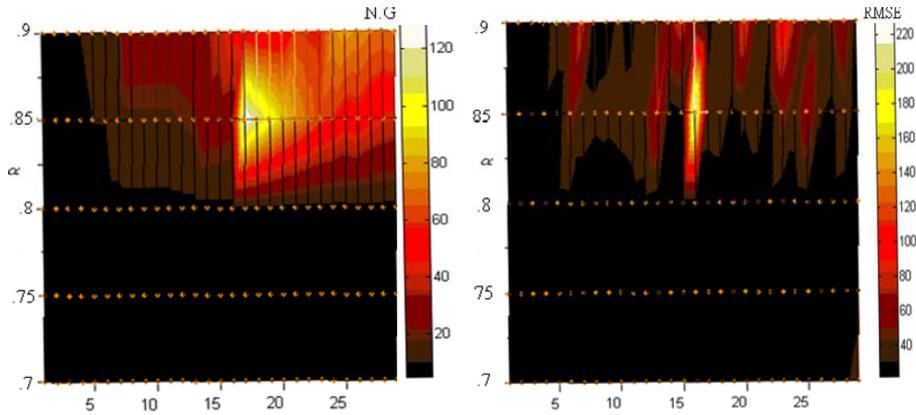

**Figure 4. Effect of $\alpha$ variations in neuron growth – $\beta$ =.001(N.G)-left- &RMSE-right- of SONFIS with n.r=2 over 30 iterations; .7=< $\alpha$ <=.9**

Let's consider a reverse state: $\alpha$ is constant (=.9) and $\beta$ takes different values (Figure 6). Such consideration, beside possible phase altering step (after $\beta = 4 \times 10^{-4}$), may display another feature of society alteration: the proper chaos related to the later state has larger values so that is not relatively agreed with N.G. In fact, our government loses pervious relative order [11]. In other process, simultaneously $\alpha$ and $\beta$ are changed over 10 iterations (Figure 7).

In a scaled coordination, it can be probed that the rate of $\alpha/\beta$ for the lugeon data set and for laminar state is near to 1, even if $\alpha$ reaches a little bigger value. This feature expresses the response of overall system to the current wave of information, is more depends on the pervious state of society than to the government reaction. Another interesting result of such accomplishing is the behavior of the system in the large values of $\alpha$ and $\beta$: despite of some anomalies, for large digits, system doesn't consider other disorder parameter and fall in to the revolutionary state.

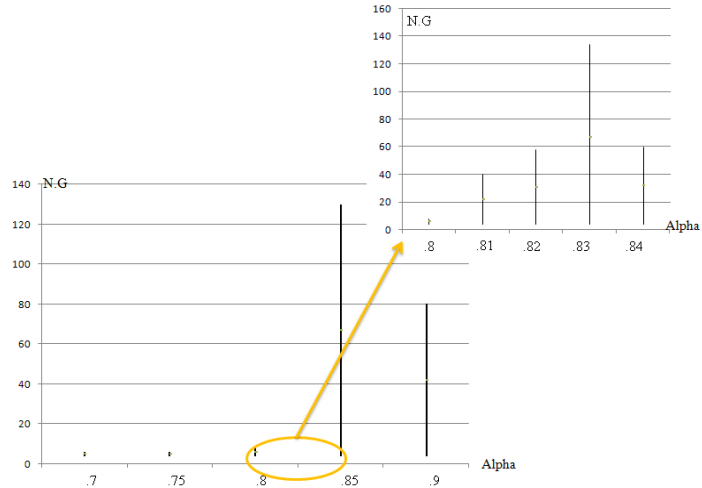

**Figure 5. Aggregation of $\alpha$ variations N.G in SONFIS with n.r=2 over 30 iterations (Phase transition step)**

Under the power law (Eq.2) and constant government noise (other effective parameters), the utilized SONFIS releases a gradual phase transition step (Figure 8). The main difference between this case and the pervious is on the emerged transition pattern at the low values of $\alpha$ which displays a sensible border proves that this system has a temporal transition under the constant circumstance and only a part of information nodes get an active role however raising of the social nodes' shadows brings up the possible oppositional disorders. In second situation we employ SORST-AS, upon this assumption that the government based on history, experience and other like fashions in the world, has ability to elicitation of relatively approximated rules of the observed and distinguished behaviors (by transferring of attributes to the changeable scaled classes using 1-D SOM, as well as low, middle, high and so on). The applied error measure for measure of performance of RST is given by:

$$MSE = \frac{\sum_{i=1}^{m}(d_i^{real} - d_i^{classified})^2}{m} \;; \qquad (4)$$

In deducing of decision for approximated rules (not unique decision part), we select highest value (largest ambiguities) for such decisions. By repeating of the pervious steps, we obtain behavior of SORST-AS where we employ $\alpha = .9$, $\beta = .7$ and $\gamma = 1$, in Eq. 1 (Figure 9). This proves how with changing of scaling, democratic government may lose the current prevalent order of the society while noise parameters keep on the initial values. In other word, more details may astound even democratic government.

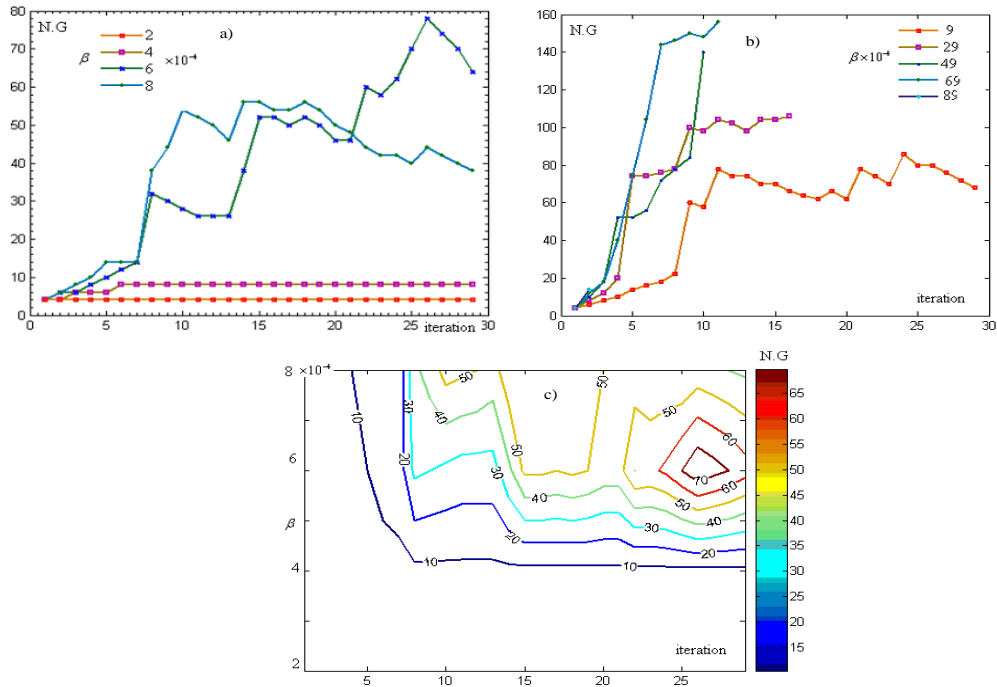

**Figure 6.** Effect of $\beta$ variations- $\alpha$ =.9 - in Neuron Growth (N.G)-left- &RMSE-right- of SONFIS with n.r=2 over 30 iterations; **a)** $2\times 10^{-4} \leq \beta \leq 8\times 10^{-4}$; **b)** $8\times 10^{-4} \leq \beta \leq 8.5\times 10^{-3}$ and **c)** Phase transition diagram contour of N.G-SONFIS

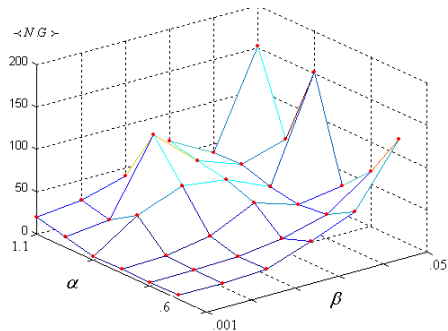

**Figure 7.** Effect of $\alpha$ & $\beta$ variations over 10 iterations on the average NG-n.r=2 (SONFIS)

## 4. Conclusion

In this study we proposed two new algorithms in which SOM, NFIS and RST make SONFIS and SORST. Mutual relations between algorithms layers identify order-disorder transferring of such systems.

So, we found our proposed methods have good ability in mimicking of government-nation interactions while government and society can take different states of responses. Developing of such intelligent hierarchical networks, investigations of their performances on the noisy information and exploration of possible relate between phase transition steps and flow of information are new interesting fields, as well in various fields of science and economy.

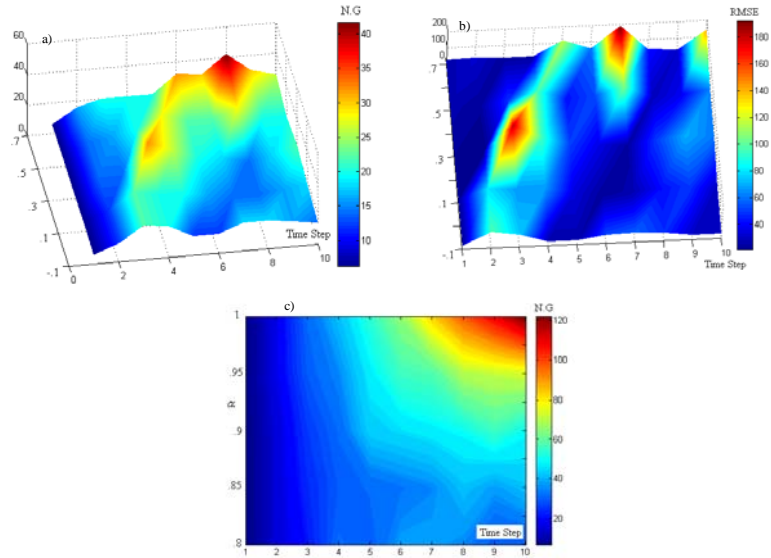

**Figure 8. Effect of $\alpha$ variations in neuron growth $\beta$ =.25, $\gamma$ =.5, SONFIS (power law) with n.r=2 over 10 iterations**

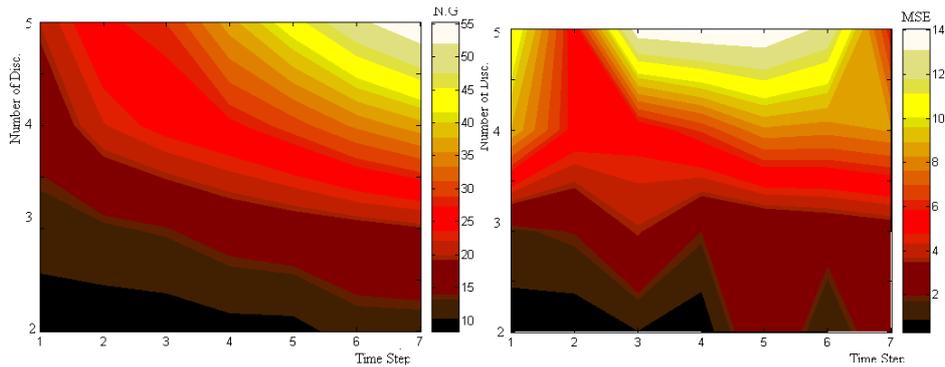

**Figure 9. Effects of scaling variations in SORST-AS on the N.G and MSE over 7 time steps-($\alpha$ =.9, $\beta$ =.7 and $\gamma$ =1)**